\documentclass[10pt,twocolumn,letterpaper]{article}

\usepackage{cvpr}              

\usepackage[accsupp]{axessibility}  

\definecolor{cvprblue}{rgb}{0.21,0.49,0.74}
\usepackage[pagebackref,breaklinks,colorlinks,allcolors=cvprblue]{hyperref}

\def\paperID{2} 
\def\confName{CVPR}
\def\confYear{2026}

\title{\methodname: Multi-person 2D-to-3D Pose Lifting}

\newcommand{\methodname}{MuPPet\xspace} 

\author{Thomas Markhorst$^1$, Zhi-Yi Lin$^1$, Jouh Yeong Chew$^2$, Jan van Gemert$^1$, Xucong Zhang$^1$\\
$^1$Delft University of Technology, $^2$Honda Research Institute Japan\\
{\tt\small t.c.markhorst@tudelft.nl}
}

\begin{document}
\maketitle
\begin{abstract}
Multi-person social interactions are inherently built on coherence and relationships among all individuals within the group, making multi-person localization and body pose estimation essential to understanding these social dynamics. One promising approach is 2D-to-3D pose lifting which provides a 3D human pose consisting of rich spatial details by building on the significant advances in 2D pose estimation. However, the existing 2D-to-3D pose lifting methods often neglect inter-person relationships or cannot handle varying group sizes, limiting their effectiveness in multi-person settings. We propose \methodname, a novel multi-person 2D-to-3D pose lifting framework that explicitly models inter-person correlations. To leverage these inter-person dependencies, our approach introduces Person Encoding to structure individual representations, Permutation Augmentation to enhance training diversity, and Dynamic Multi-Person Attention to adaptively model correlations between individuals. Extensive experiments on group interaction datasets demonstrate \methodname significantly outperforms state-of-the-art single- and multi-person 2D-to-3D pose lifting methods, and improves robustness in occlusion scenarios. Our findings highlight the importance of modeling inter-person correlations, paving the way for accurate and socially-aware 3D pose estimation.
Our code is available at: \href{https://github.com/Thomas-Markhorst/MuPPet}{https://github.com/Thomas-Markhorst/MuPPet}

\end{abstract}    
\section{Introduction}
\label{sec:intro} 
Nonverbal cues such as body pose are closely linked to group dynamics \cite{ennis2010seeing} and internal human states, including intention and emotion \cite{pose_emotion_2023}. These cues also play a key role in effective human communication \cite{lee2019talking,marcos2013body,hall2019nonverbal}.
Therefore, accurately detecting the human pose of each person within a multi-person interaction is crucial for interpreting social cues. It has been shown that in a group activity, the head and body orientation \cite{varadarajan2018joint,alameda2015analyzing}, individual body actions \cite{balazia2022bodily}, and body pose associated with facial expression \cite{metaxas2013review}, are important for understanding social interaction and group dynamics. 
In addition, the perception of multi-person body motion is desirable for intelligent systems such as humanoid robots to interact with groups~\cite{bennewitz2005towards}.
Noticeably, 3D human pose in absolute space is preferred over relative space in multi-person pose estimation since absolute pose can relate the positions and orientations
between individuals in the real-world space, which is important for social interaction analysis \cite{varadarajan2018joint,alameda2015analyzing}.

\begin{figure}[t]
    \centering
    \includegraphics[width=\columnwidth]{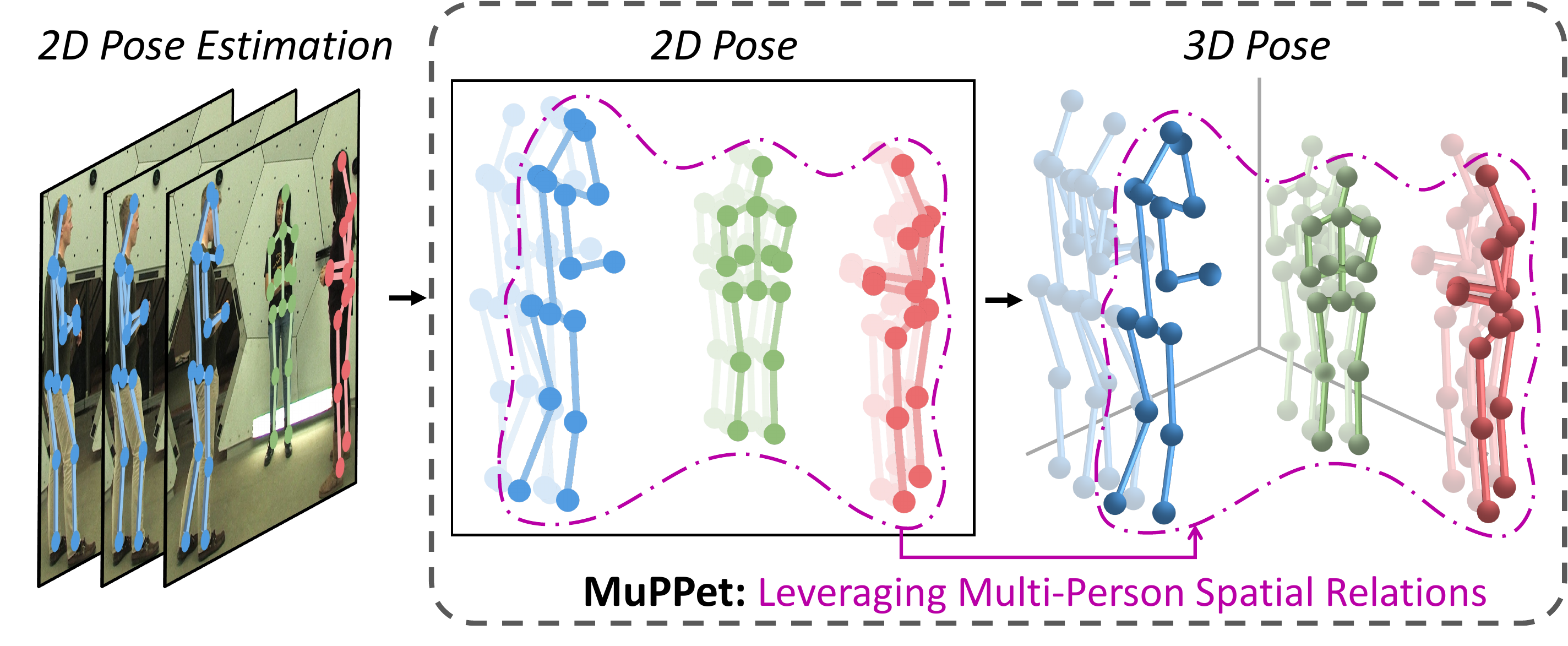}
    \caption{
     We exploit social inter-person correlations for 3D pose estimation to help infer occluded poses and spatial relations between individuals in a group interaction. Our method takes a sequence of detected 2D body poses to predict the sequence of 3D poses, as shown in the figure. 
    }
    \label{fig:teaser}
\end{figure}

Despite the blooming development of 3D human pose estimation \cite{toshev2014deeppose,zhang2019fast,peng2024dual,mehraban2024motionagformer}, most existing works neglect the correlations between persons in social interaction.
It is noteworthy that the correlations of multi-person body movements during group social interactions have been widely discussed as ``imitation'', ``mimicry'', ``synchrony'' \cite{correlation2024}.
It is promising to leverage such interactions between individuals in a group to enhance pose estimation performance in multi-person scenes.
Unfortunately, most 3D pose estimation methods focus on single-person scenarios \cite{wang2021deep,zheng2023deep,munea2020progress}.
Although there are previous studies on multi-person 3D pose estimation \cite{su_virtualpose_2022,ROMP,sun_putting_2022}, they mainly consider the crowded setting in terms of occlusion or depth placement and ignore any interaction between people. 
POTR-3D \cite{park_towards_2023} is an exception as they model inter-person interaction. However, this method is trained with fixed group size \cite{park_towards_2023}, which is impractical in real-world scenarios.

In this paper, we propose a 2D-to-3D body pose lifting method \methodname specifically designed for the multi-person setting, which models the correlation information between persons via spatial attention, person encoding, and permutation learning.
We pick the 2D-to-3D pose lifting approach due to its promises of leveraging accurate 2D pose estimation and focusing on the significance of the spatial 3D pose details \cite{martinez2017simple,zhang_mixste_2022,park_towards_2023,qian_hstformer_2023}.
An overview of \methodname is shown in Fig.~\ref{fig:teaser}, which takes the detected sequence of 2D poses as input and lifts them to the absolute 3D body pose in the world coordinate system. 

Extending from the typical self-attention module in a single person \cite{zhang_mixste_2022}, we apply the self-attention on the collection of persons to capture the relationship between people.
We incorporate a person encoding into our model to associate each body joint with their corresponding individual explicitly, which causes the model to efficiently learn inter-person relationships. Building on this person encoding, we introduce a novel data augmentation strategy that permutes person ordering during training to enhance training diversity. 
The combination of spatial attention and person encoding enables \methodname to handle arbitrary numbers of people in the scene, instead of a fixed maximum number of people~\cite{park_towards_2023}.
We implement our approach using a diffusion process with a transformer backbone and include the temporal information across frames in a video.

Through extensive experiments, we demonstrate that the proposed method not only outperforms the single-person baseline but also exceeds the performance of current state-of-the-art 2D-to-3D multi-person lifting methods. 
Notably, our method exhibits clear advantages in handling occlusions compared to the single-person method. 
We believe this approach represents an important utility for advancing human behavior analysis in multi-person settings.
In summary, our contributions are three-fold:
\begin{itemize}
    \item A method for pose lifting a dynamic number of persons in a group interaction across frames to 3D.
    \item Efficient intra- and inter-person modeling with proposed person encoding, permutation learning, and spatial attention across all persons.
    \item Better performance and occlusion handling compared with a single-person baseline, and outperforms state-of-the-art (SOTA).
\end{itemize}
 \section{Related work}
\label{sec:related_work}
\paragraph{3D Pose Estimation.}
Estimation of a 3D human pose from a 2D image is a long-standing challenging problem \cite{wang2021deep,sarafianos20163d}.
A straightforward solution for this task is directly taking the input to predict 3D pose \cite{zhen_smap_2020,moon_camera_2019,mehta2017vnect,chen20173d,ci2019optimizing,gong2021poseaug}. Prior knowledge of human kinematics has been investigated in previous works \cite{peng2024ktpformer}.
Especially, the SMPL body model \cite{SMPL:2015} is commonly used in these 3D pose estimations \cite{sun_putting_2022,yuan2021simpoe,song2020human,zhang2024dynamic}.
While these approaches can handle a variety of scenes, they require accurate 3D annotations, which is expensive and labor-intensive. 
In contrast, other methods \cite{martinez2017simple,zhang_mixste_2022, park_towards_2023,qian_hstformer_2023} lift 2D pose detections produced by 2D pose estimators trained on large datasets to 3D pose. These lifting methods reduce the need for 3D annotations in diverse environments and have recently gained popularity. 
Various architectures have been explored for single-person lifting, including transformers \cite{zhang_mixste_2022,li_mhformer_2022,qian_hstformer_2023,mehraban_motionagformer_2024,li2024hourglass,mehraban2024motionagformer}, fully connected networks \cite{ci2022gfposelearning3dhuman}, graph convolutional networks \cite{GCN_lifting, Zou_2021_ICCV}, and diffusion models \cite{shan_diffusion-based_2023, choi_diffupose_2023,xu2024finepose,jiang_back_2024}. 
However, due to the limited 3D annotated data, data augmentation is commonly used and shows effectiveness for the pose lifting task \cite{peng2024dual,park_towards_2023}.
In this paper, we focus on 2D-to-3D lifting and propose a diffusion-based approach with a transformer-based denoiser.

A second distinction in 3D pose estimation is the use of temporal information \cite{hossain2018exploiting}. Single-frame methods (frame2frame) \cite{poselifter} rely solely on per-frame inputs, making them sensitive to occlusions and noise when applied to a video.
Sequence-to-frame methods \cite{pavllo20193dhumanposeestimation, Liu_2020_CVPR} improve robustness by predicting pose on one frame using a sequence of past and future frames. 
Moreover, sequence-to-sequence models \cite{zhang_mixste_2022, tang_3d_2023, zhu_motionbert_2023,zheng20213d,zhao2023poseformerv2} enforce temporal consistency across frames and enable more efficient inference. Based on the temporal handling advantages, we incorporate both input and output sequences in our approach.

\label{sec:method}
\begin{figure*}[htbp]
    \centering
    \includegraphics[width=0.95\textwidth]{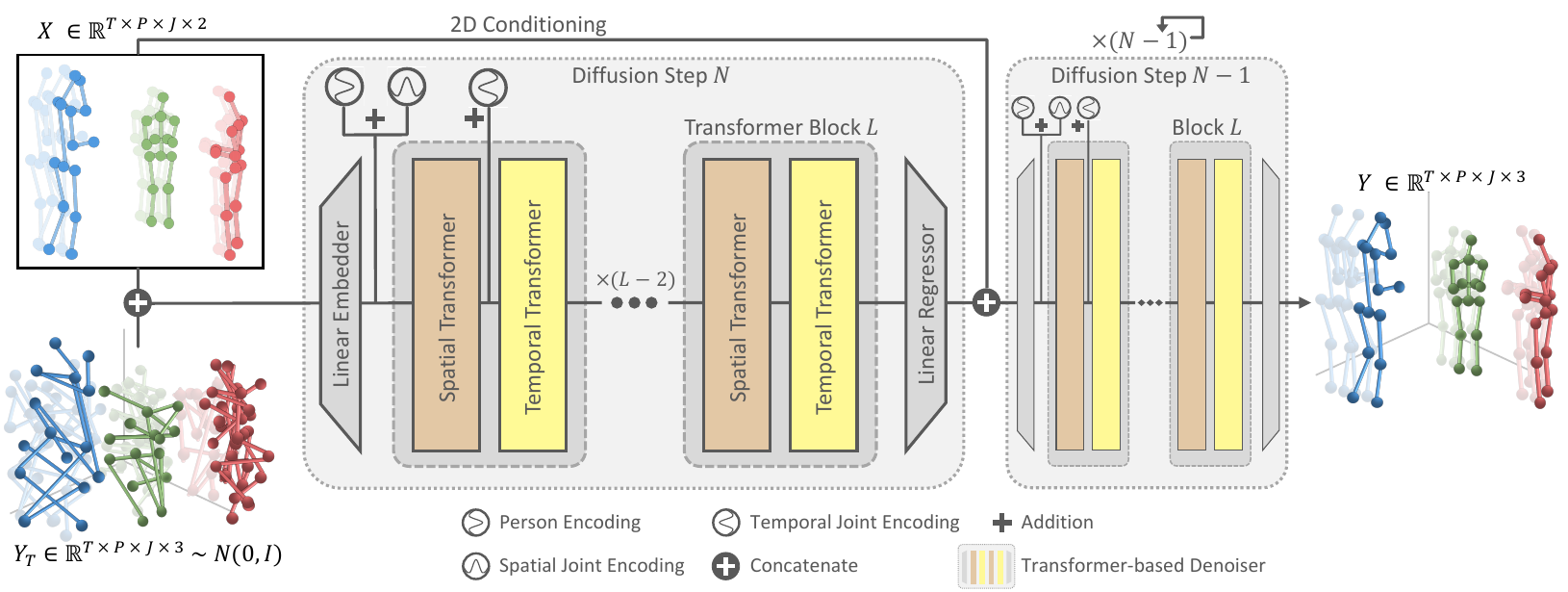}
    \caption{Overview of our \methodname pipeline. Given a sequence of detected 2D human pose joints from multiple persons \(X\), we use the diffusion process \(N\) times to denoise the 3D random poses \(Y_N\) to the output absolute 3D pose \(Y\). Inside the denoiser, the spatial transformer and person encoding are applied to capture intra- and inter-person relationships, and a temporal transformer is used to capture the joint relationship across frames.}
    \label{fig:pipeline}
\end{figure*}

\paragraph{Multi-person Pose.}
Multi-person 3D pose estimation introduces two primary challenges: placing poses in an absolute space and handling occlusions caused by other individuals. Direct 3D methods, such as BEV \cite{sun_putting_2022}, excel at absolute positioning by explicitly modeling a bird’s-eye-view representation. ROMP \cite{ROMP} incorporates a collision-aware strategy to mitigate inter-person overlap, while 3DCrowdNet \cite{3DCrowdNet} enhances robustness in crowded scenes by integrating an additional 2D pose extractor. However, these methods are frame-to-frame due to the high computational cost of end-to-end video processing. Moreover, they require extensive diverse training data to cover variances in appearance. 
To reduce dependency on diverse 3D training data, 2D-to-3D lifting approaches have also been explored. VirtualPose \cite{su_virtualpose_2022} predicts 3D pose using detected bounding boxes and a joint heat map, removing the image input from the 3D estimation. 
In contrast, POTR-3D \cite{park_towards_2023} handles occlusion better due to its sequence-to-sequence modeling and leveraging of ground-truth data during training. Unfortunately, it is trained with fixed numbers of people in a scene and pads less crowded frames with zeros. Additionally, POTR-3D~\cite{park_towards_2023} applies data augmentations that separately translate and rotate individual persons. Such data augmentations bring performance improvements, however, they disrupt the inter-person relationships.
To form a multi-person augmentation leveraging inter-person relationship, we propose the person encoding and then permute the person ordering to simulate various multi-person settings.

\paragraph{Diffusion.}
Diffusion models, first introduced by \cite{diffusion_original} and specifically the Denoising Diffusion Probabilistic Models (DDPM) \cite{DDPM}, have demonstrated strong performance across various generative tasks \cite{rombach_high-resolution_2022, diff1, diff2, diff3, diff4, diff5}. 
Their probabilistic nature makes them well-suited for addressing key challenges in 3D human pose estimation such as occlusion and depth ambiguity. 
Given any 2D input, deterministic models must commit to a single prediction, potentially losing plausible alternatives, especially in occluded cases. 
In contrast, diffusion models can probabilistically reason over multiple valid 3D poses, generating diverse hypotheses instead of making a fixed choice. 
This has led to a few single-person pose-lifting methods using diffusion \cite{zhang_mixste_2022,shan_diffusion-based_2023}.
These methods generate hypotheses, which can then be aggregated to refine final pose predictions and improve overall performance \cite{shan_diffusion-based_2023, holmquist_diffpose_2023}, making diffusion-based approaches a promising direction for 3D pose estimation.

\section{Method}

Our target scenario is a multi-person social setting where the individuals are engaged in one activity, for example, a group dialogue.
\methodname takes a sequence of 2D body poses from multiple persons to output the lifted 3D human poses of all individuals in the sequence.

\subsection{Architecture}
\label{subsec:architecture}
The overview of the proposed method is shown in Fig.~\ref{fig:pipeline}.
Our method takes a detected sequence of 2D body poses \(X \in \mathbb{R}^{T \times P \times J \times 2}\) of \(T\) frames, with \(P\) persons, \(J\) joints, and 2D joint positions as input. 
A linear layer maps \(X\) to an embedding \(\hat{X} \in \mathbb{R}^{T \times P \times J \times 512}\), which is fed into an attention block with spatial and temporal attention modules. The attention block is repeated $L$ times to generate the features that are fed into a regression head that maps the final attention output \(\hat{X_L}\) to 3D multi-person pose \(Y \in \mathbb{R}^{T \times P \times J \times 3}\).
We employ the diffusion process similar to previous works \cite{shan_diffusion-based_2023,choi_diffupose_2023,xu2024finepose,jiang_back_2024} to gradually generate the 3D pose with \(N\) times denoising process.

\paragraph{Multi-person spatial attention.} 
Inspired by \cite{zhang_mixste_2022} that performs the attention within a single person, we extend it to capture all joints across any number of persons in the scene. Given the encoded feature from a frame $t$ as \(f_t \in \mathbb{R}^{P \times J \times 512}\), we reshape the representation to \(\hat{f}_t \in \mathbb{R}^{(P \cdot J) \times 512}\)
, where \(P\) is a variable to handle varying group sizes. The query, key and value matrix \(Q, K, V\) are computed using collection of tokens \(\hat{f}_t\) and weights \(W^K, W^Q, W^V\). Followed by a linear projection \(W^L\). Note all learned weights \(W^K, W^Q, W^V, W^L\) have dimension \(\mathbb{R}^{512 \times 512}\) which is independent of the amount of persons \(P\) and can therefore handle dynamic group sizes. In this way, we manage to apply self-attention across joints from all persons in the scene. To enable the model to handle varying numbers of persons, the model is trained with varying $P$, ensuring that both inter-person and intra-person spatial relations are captured for all group sizes.

\paragraph{Temporal attention.} 
As our method is sequence-to-sequence, it should relate each joint across all frames. To achieve this, we first collect a sequence of tokens \(s_{p,j} \in \mathbb{R}^{T \times 512}\). We then perform self-attention with \(s_{p,j}\) with a temporal embedding similar to \cite{zhang_mixste_2022}. The query, key and value matrix \(Q, K, V\) are computed using collection of tokens \(\hat{s}_{p,j}\) and weights \(W^K, W^Q, W^V\). Followed by a linear projection \(W^L\).
Following the attention operation, the representations $s_{p,j}$ for all persons $p$ and joints $j$ are concatenated and reshaped to form $\hat{X}_l \in \mathbb{R}^{T \times P \times J \times 512}$, which serves as the output of the spatio-temporal block.

\subsection{Person Encoding}
Both the spatial and temporal transformer modules do not distinguish which person a joint belongs to.
This introduces challenges to using the inter-person information for 3D pose lifting. To explicitly encode the person information into the model, we propose the person encoding \(E \in \mathbb{R}^{P \times 512}\), which is the same for all joints \(j \in J_p\) that belong to person \(p\).
$E$ is a set of learned parameters.
The person encoding ensures the model distinguishes the joints that belong to individuals while considering the relationship between people.
It addresses the issue in previous multi-person pose lifting \cite{park_towards_2023} that can only process the fixed maximum number of persons and add zero padding to missing persons.
Together with the spatial attention across all joints, the person position embedding grants the capability of our method to handle multi-person sequences and, importantly, dynamic group sizes. 
Specifically, $E$ is added to $\hat{X}$ before the first spatial transformer block as $\hat{X}_{p,j} = \hat{X}_{p,j} + E_p$ for all $p \in P$ and $j \in J$.
Similar to the spatial and temporal joint encoding, $E$ is added to $\hat{X}$ again at every diffusion timestep. 
 
\begin{figure}[t]
    \centering
    \includegraphics[width=0.93\columnwidth]{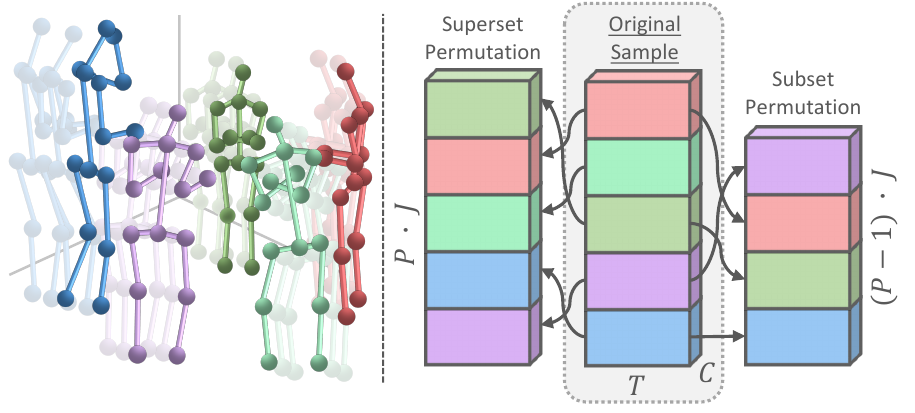}
    \caption{Example of permutation learning for a five person scene. We illustrate the people with color coding on the left and the permutation in the feature space on the right. From the original feature $\hat{X} \in \mathbb{R}^{T \times (P \cdot J) \times C}$, where \(C\) is the number of channels, we show a possible superset permutation of \(P\) persons with all \(J\) joints, and subset permutation of \((P-1)\) persons with all \(J\) joints.
    During training, $n_\text{sub}$ and $n_\text{sup}$ permutations are used.
    }
    \label{fig:permutation}
\end{figure}

\subsection{Permutation Learning}
As the multi-person pose lifting requires the consideration of the relationship between people, the commonly used data augmentations in the single-person setting, such as translation and rotation, cannot be applied. Interestingly, our newly proposed person encoding allows us to perform a new permutation data augmentation specifically for the multi-person 3D pose lifting task.

Given an input-output pair \(X, Y\) which consists of \(P\) persons, we highlight that there is no specific ordering of the individual persons in $P$. Consequently, arbitrarily permuting the order of \(P\) fed into the model could increase the diversity of the data. However, the spatial and temporal attention modules themselves are invariant to such person order permutations since they do not distinguish individuals. 
Fortunately, our person encoding explicitly assigns the embedding $E_p$ to each person $p$ in the scene to enable the model to distinguish individuals between permutations. We exploit this by randomly permuting the order of \(P\) of any sample \(X\), called \textit{superset permutation}. The person encodings are added to samples only after permuting. 
We illustrate the permutation learning with an example in Fig.~\ref{fig:permutation}.

Additionally, we hypothesize that a subset of \(P\) in a social setting exhibits similar relations as its superset. Therefore, we also take varying subsets of \(P\) and randomly permute them, which we call \textit{subset permutation}. Taking both strategies together, we result in the permutation learning approach where for each sample \(X\) we train on i) \(X\) itself in normal person order, ii) \(n_{\text{sup}}\) random permutations of the person order of \(X\) and iii) \(n_{\text{sub}}\) evenly divided over the two subset sizes \(|P| - 1\) and \(|P| - 2\).
We define \(n_{sup}\) as the number of permutations of the superset, and \(n_{\text{sub}}\) as the number of permutations for the subset.

\subsection{Diffusion Process}
Following DDPM \cite{DDPM}, we employ a diffusion process to gradually corrupt our multi-person 3D pose \(Y \in \mathbb{R}^{T \times P \times J \times 3}\) by adding Gaussian noise over \(N\) time steps, such that \(Y_N\) becomes nearly pure Gaussian noise. The transformer architecture described in Section \ref{subsec:architecture} is used as a denoiser in the reverse diffusion process. The process is defined in Eq.~\ref{eq:forward_diffusion}, where \(\bar{\alpha}_t\ \in [0,1]\) is a fixed hyperparameter to control the noising scheme and $Y_0=Y$:

\begin{equation}
    Y_n = \sqrt{\bar{\alpha}_n}Y_{0} + \sqrt{1 - \bar{\alpha}_n}\epsilon, \quad\epsilon \sim \mathcal{N}(0, I) 
    \label{eq:forward_diffusion}
\end{equation}

We condition the reverse diffusion process on detected 2D pose \(X\). Following \cite{shan_diffusion-based_2023}, we concatenate \(X\) with \(Y_n\) to form \(Z_n \in \mathbb{R}^{T \times P \times J \times 5}\) at diffusion step $n$. From every \(Z_n\) the denoiser outputs \(\tilde{Y_0}\) targeting \(Y_0\), following DDIM \cite{DDIM}. With any \(n' < n\), \(Y_{n'}\) is constructed by adding noise as shown in Eq.~\ref{eq:backward_diffusion}.
Then \(Y_{n'}\) is again concatenated with \(X\) to construct \(Z_{n'}\).

\begin{equation}
    Y_{n'} = \sqrt{\bar{\alpha}_{n'}}\cdot\tilde{Y_0} + \sqrt{1 - \bar{\alpha_{n'}}}\cdot\epsilon + \sigma_n\epsilon
    \label{eq:backward_diffusion}
\end{equation}
\begin{equation}
    \tilde{Y_0}=\text{denoise}(\text{concat}(X,Y_n))
\end{equation}
 
At inference, we sample multiple \(Y_N\) conditioning on the same \(X\), which gives multiple outputs. 
We then aggregate these by taking the joint-wise average, to boost performance as in previous works \cite{shan_diffusion-based_2023,holmquist_diffpose_2023}. 

\subsection{Pose Loss} 
We experimentally find that directly outputting absolute positions of all joints is a difficult task. Therefore, instead of predicting all joints of \(Y\) in absolute space, we normalize poses by separating relative pose and absolute root location. For the relative pose, each joint is transformed into a root-relative coordinate system, where the hip-center serves as the root joint, following standard practice \cite{zhang_mixste_2022, tanke_social_2023}. 
The root-relative positions typically fall within \([-1, 1]\) meters, requiring no further normalization. However, the absolute root joint is normalized based on the training set as it has a larger range.
The single absolute root joint and remaining relative joints have separate loss terms \(\mathcal{L}_{\text{abs}}\) and \(\mathcal{L}_{\text{rel}}\), respectively. To balance the weight between relative and absolute joint calculation, we introduce a weight $\lambda$:

\begin{equation}
    \mathcal{L}_{\text{MPJPE}} = \lambda \cdot \mathcal{L}_{\text{abs}} + (1 - \lambda) \cdot \mathcal{L}_{\text{rel}}
    \label{eq:loss}
\end{equation}

\section{Experiments}
\label{sec:experiments}
We first compare our multi-person method \methodname with the SOTA single-person 3D pose lifting method to highlight its advantages in the multi-person setting. 
We then compare \methodname to the current SOTA multi-person 3D pose lifting methods with the reported numbers in their original papers. We further show the effectiveness of \methodname in handling occlusions.
Finally, we conduct ablation studies to analyze the impact of individual modules in our proposed method.

\subsection{Experimental Settings}
\paragraph{Datasets} 
The Haggling dataset \cite{joo_towards_2019} contains 30 recordings with 173 separate sequences inside a capture system with 31 cameras. Within each recording, three participants play a haggling activity for around one minute. Each person is annotated with 19 body pose joints following the COCO19 format. It captures social interaction \cite{tanke_social_2023} due to participants having specific social roles compared to other datasets \cite{mehta_single-shot_2018,Marcard_2018_ECCV}, which increases its relevance for our study.
Models evaluated on the Haggling dataset are trained on 133 sequences, the corresponding test set contains 40 sequences. We use six superset and six subset permutations due to having only three persons in each scene.

CMU Panoptic \cite{panoptic_joo} is an older dataset than Haggling \cite{joo_towards_2019} and is recorded in the same capture system. It contains several social games varying in group size from two to eight persons with 13 recordings and 58 separate sequences, significantly less compared to Haggling \cite{joo_towards_2019}. Different from the 19 body joints annotated in the Haggling dataset, Panoptic has 15 body joints following the MPI15 format. 
Following \cite{park_towards_2023,su_virtualpose_2022,zhen_smap_2020}, we train and evaluate on cameras with the index of 16 and 30.
We focus on the Haggling, Mafia, and Ultimatum scenes in the dataset, as these sessions contain a single group interaction. Other scenes, like Pizza, contain either fragmented groups or single persons and are therefore not suitable for our group based lifting method.
The model evaluated on Panoptic is trained on 38 sequences and tested on 20 sequences. Due to the low number of training samples, models tested on Panoptic are pre-trained on Haggling and finetuned with half the initial learning rate. Due to the small size of this dataset and having more subjects compared to the Haggling dataset, we set the permutation to be 13 superset and three subset permutations.

\textbf{Implementation Details.} We set \(L=8\) such that the network contains eight spatio-temporal blocks.
The network is optimized for 400 epochs using AdamW \cite{loshchilovdecoupled} in PyTorch, with a starting learning rate of 6E-5 and exponential decay of 0.997 per epoch. The maximum timesteps for the diffusion process are set to 1000, the batch size is four, and we train on pose sequences of 243 frames following \cite{zhang_mixste_2022}. Training and inference are run on a single NVIDIA A100 GPU. For all models, the input 2D pose sequences are obtained by OpenPose detection \cite{openpose}, and person-id is matched with ground-truth.

\textbf{Metrics.} Following \cite{tanke_social_2023} for the Haggling dataset, and \cite{park_towards_2023, sun_putting_2022, su_virtualpose_2022} for the CMU Panoptic dataset, we report mean per joint positional error (MPJPE) in relative space and absolute space with mm as a unit.
We only consider joints that are inside the camera frame. For the relative joint error calculation, we align the predicted and GT root joint and average the error.
For the absolute joint error calculation, we take the global origin and calculate the average joint error relative to the origin. The root error is the error in absolute space of only the root joint estimation.

\subsection{Comparison of Multi-person and Single-person}

To reveal the benefit of correlation learning between persons, we conduct experiments on the Haggling dataset due to the rich interactions between people.
We pick the SOTA single-person 3D pose lifting method D3DP \cite{shan_diffusion-based_2023} as the baseline in this experiment, which is similar to our method in terms of using diffusion for human pose lifting, while only processing a single person individually. 
We only compare the D3DP since it shows superior performances compared with current other single-person 3D pose lifting methods in the original paper \cite{shan_diffusion-based_2023}.
The original D3DP is trained only on the relative joints, we alter it to predict both relative and absolute joint positions (D3DP$_\text{absolute}$). Note that we evaluate D3DP with the exact same training and test sets as \methodname for a fair comparison.

In Tab. \ref{tab:singlevsours}, we can see that \methodname outperforms the D3DP baseline in both relative and absolute joint position errors, with significant margins of 6.4\% (from 59.1 to 55.3) in the relative, 17.2\% (from 135.9 to 108.9) in the absolute joint position errors, and 19.9\% in the absolute root joint error. It clearly demonstrates the benefits of leveraging intra- and inter-person relationships for the 3D pose-lifting task.
The improvement in absolute joint error is expected as we hypothesized that modeling multi-person relations gives an improved understanding of 3D location. 
We attribute the improvement in relative joint error to our permutation learning approach used as data augmentations, which cannot be applied to single-person methods.
Note that the performance improvement of the absolute joint error is higher than the relative (17.2\% vs 6.4\%), which strongly indicates the benefit of \methodname in handling the absolute joint estimation. 
This is relevant, as the absolute joint position is more meaningful than the relative joint position in the multi-person interaction to understand the group dynamics.

By comparing the original D3DP and our alternation with absolute joint loss, we can see that adding the absolute joint loss slightly decreases the relative pose performance. Which could be caused by predicting relative and absolute joint positions jointly being a difficult task.
\begin{table}[t]
\centering
\resizebox{\columnwidth}{!}{%
\begin{tabular}{@{}lccc@{}}
\toprule
Method                                                                   & MPJPE$_{\text{rel}}$ $\downarrow$ & MPJPE$_{\text{abs}}$ $\downarrow$ & MPJPE$_{\text{root}}$ $\downarrow$ \\ \midrule
D3DP \cite{shan_diffusion-based_2023}                                    & 58.2                     & -                        & -                         \\
D3DP$_{\text{absolute}}$ \cite{shan_diffusion-based_2023} & 59.1                     & 144.4                    & 135.9                     \\
\methodname                                                             & \textbf{55.3}                     & \textbf{119.5}                   & \textbf{108.9}                     \\ \bottomrule
\end{tabular}%
}
\caption{Comparison of \methodname with single-person lifting method D3DP on the Haggling dataset, in absolute (MPJPE$_{\text{abs}}$), relative (MPJPE$_{\text{rel}}$), and absolute root MPJPE$_{\text{root}}$ pose estimation in mm. Our \methodname achieves better performance than the SOTA single-person pose lifting method D3DP.}
\label{tab:singlevsours}
\end{table}

\subsection{Comparison with SOTA} 
It is difficult to compare with the previous SOTA methods due to the limited datasets with group interaction, availability of source code from previous works, and the lack of absolute joint error metric.
In this section, we make an effort to compare \methodname with the current SOTA 3D pose estimation method including 2D-to-3D pose lifting and direct 3D pose estimation.
We use Panoptic in this experiment as it is popular for the 3D pose lifting task.
We report only the relative joint error due to missing absolute joint error results. We focus on the Haggling, Mafia, and Ultimatum scenes in Panoptic, as they fit our target setting where all people engage in one group activity.
Since \methodname is a 2D-to-3D multi-person pose lifting method, we mainly compare it against VirtualPose \cite{su_virtualpose_2022} and POTR-3D \cite{park_towards_2023}.
We also list the performances reported by other SOTA multi-person direct 3D pose estimation models~\cite{zhen_smap_2020, MUBYNet}.

The results are shown in Tab. \ref{tab:ours_vs_sota}, with performances in individual scenes and averaged errors across Haggling, Mafia, and Ultimatum scenes. 
We can see from the table that multi-person 2D-to-3D pose lifting methods, \ie VirtualPose \cite{su_virtualpose_2022}, POTR-3D \cite{park_towards_2023}, and our \methodname, achieve better performances than direct estimation methods. 
Among the top three methods, our \methodname achieves the best performance in terms of the averaged relative joint errors across the three scenes.
Note that, although the optimization target of our method is the absolute joint pose, \methodname still outperforms previous SOTA 2D-to-3D methods in the relative joint error. 
It shows the benefit of leveraging the intra- and inter-person relationships for multi-person pose estimation.

\begin{table}[t]
\resizebox{\columnwidth}{!}{%
\begin{tabular}{@{}lcccc@{}}
\toprule
Method                                 & Haggling & Mafia & Ultimatum & \textbf{Mean}\\ 
\midrule
MubyNet \cite{MUBYNet}                 & 72.4     & 78.8  & 66.8      & 72.7  \\
SMAP \cite{zhen_smap_2020}             & 63.1     & 60.3  & 56.6      & 60.0  \\ 
VirtualPose \cite{su_virtualpose_2022} & \textbf{54.1}     & 61.6  & \textbf{54.6}      & 56.8 \\
POTR-3D \cite{park_towards_2023}       & 60.0     & 57.0  & 55.5      & 57.5  \\
\midrule
\methodname                          & 56.1     & \textbf{54.3}  & 57.1      & \textbf{55.8} \\ \bottomrule
\end{tabular}%
}
\caption{Comparing our method with SOTA Multi-Person pose lifting (VirtualPose and POTR-3D) and direct estimation methods (MubyNet and SMAP) on the on the CMU Panoptic dataset. All numbers are averaged joint error in mm. We only list the relative joint error \(\text{MPJPE}_{\text{rel}}\) since the absolute joint error \(\text{MPJPE}_{\text{abs}}\) is not reported by previous pose lifting methods. Our \methodname achieves the best performance in the multi-person pose estimation setting.}
\label{tab:ours_vs_sota}
\end{table}

\begin{figure*}[htbp]
    \centering
    \includegraphics[width=0.98\textwidth]{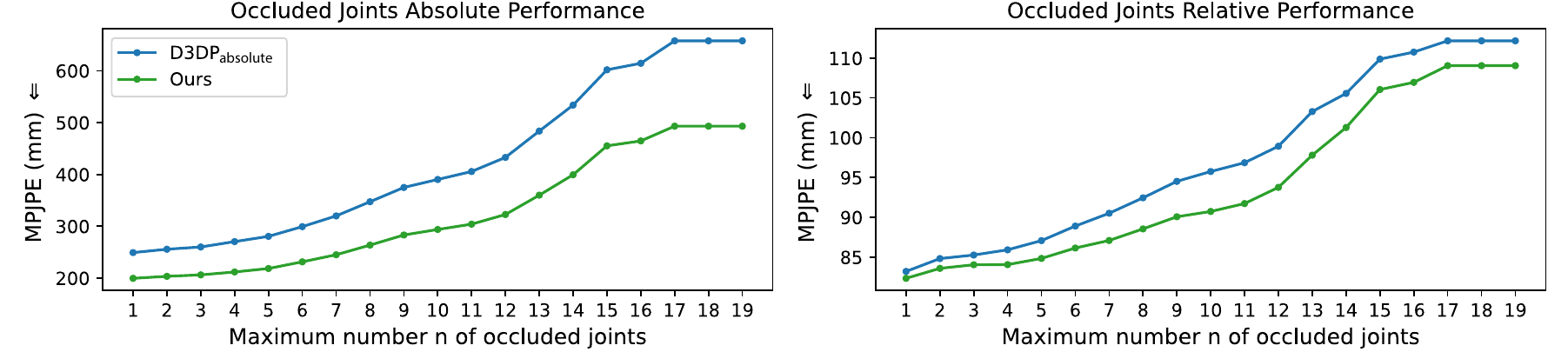}
    \caption{We show the performance of different levels of occlusion on the Haggling dataset, in comparison with the adapted D3DP singleperson model. The X-axis is the maximum number of occluded joints, and the Y-axis is the joint error in mm. Our method performs better in relative joint error MPJPE\(_{\text{rel}}\) and signficantly better in absolute joint error MPJPE\(_{\text{abs}}\) over all levels of occlusion.}
    \label{fig:exp_occlusion}
\end{figure*}

\subsection{Occlusion study}
In social interactions with multi-people, individuals often occlude each other, making pose estimation challenging. Unlike the single-person pose estimation method, our multi-person pose estimation approach can infer the occluded body pose from surrounding other people.
We assume that social interactions exhibit coherence between individuals, and our method exploits it to improve pose estimation under occlusions. 
To validate this, we compare our method with the single-person baseline D3DP on the Haggling dataset, which has rich interaction among people. We summarize the results in the Fig.~\ref{fig:exp_occlusion} with the relative joint error and absolute joint error computed over different numbers \(n\) of occluded joints per person, where $n$ is the maximum number of occluded joints for one person.

The figure clearly shows that our method handles occlusion better than D3DP across any number of occluded joints. 
The performance gap becomes larger as the number of occluded joints increases, with the effect being particularly pronounced in the absolute joint error.
Since we focus on improving the absolute joint estimation in multi-person settings, which are often occluded, the result indicates the effectiveness of \methodname.

\subsection{Ablation study}
\label{subsec:exp_ablation}
\paragraph{Component Ablation}
In this section, we examine the different components of the proposed method and their effect, using the Haggling dataset. We conduct ablation studies on dynamic multi-person handling, person encoding, permutation learning, and subset permutation learning. In Tab.~\ref{tab:module_ablation}, we show the performances by gradually adding each component to our model.
The first row in the table is the single-person baseline. By integrating our multi-person architecture, the absolute joint performance increased (from 144.4 to 142.1), however, at the cost of relative joint performance (from 59.1 to 68.4). Adding person encoding significantly improves the absolute joint performance to 131.5 while the relative joint performance gets slightly worse (from 68.4 to 69.1). Combined with superset permutation learning, absolute joint error decreases to 126.9 and relative joint estimation performance gets on par with the single-person baseline (60.0 vs 59.1). Finally, adding subset permutation learning decreases relative joint error to 55.3, and also decreases the absolute joint error such that our final model outperforms its other version by more than 5.8\% (from 126.9 to 119.5). 

\begin{table}[t]
\resizebox{\columnwidth}{!}{%
\begin{tabular}{@{}cccccc@{}}
\toprule
Multi       & PE         & Sup.Perm.      & Sub.Perm. & MPJPE$_\text{rel}\downarrow$ & MPJPE$_\text{abs}\downarrow$ \\ \midrule
-           & -          & -          & -          & 59.1                    & 144.4                   \\
\checkmark  & -          & -          & -          & 68.4                    & 142.1                   \\
\checkmark  & \checkmark & -          & -          & 69.1                    & 131.5                  \\
\checkmark  & \checkmark & \checkmark & -          & 60.0                    & 126.9                  \\
\checkmark  & \checkmark & \checkmark & \checkmark & \textbf{55.3}                & \textbf{119.5}                \\ \bottomrule
\end{tabular}%
}
\caption{Ablation showing that each module contributes to the performance. Dynamic multi-person attention (Multi), Person Encoding (PE), super set permutation (Sup.Perm, and subset permutation learning (Sub.Perm). We show both relative joint error MPJPE$_{rel}$ and absolute joint error MPJPE$_{abs}$.
}
\label{tab:module_ablation}
\end{table}

\paragraph{Permutation ablation} 
We propose two permutations, the superset and subset permutations, to improve the diversity of the training samples. To determine the effect of permutation learning and find the optimal number of permutations during training, we perform a study on different settings of permutations.
We perform this ablation on the Panoptic dataset since it contains a various number of subjects to test the subset permutations.
We report the joint error averaged over all four scenes.
First, we test the balance between superset and subset permutations. 
We keep the total number of permutations as 12 due to the trade-off between performance and computation, and change the ratio of superset and subset permutations. We show the results in the upper part of Tab.~\ref{tab:permutation_ablation}, which indicate that using subset permutations has a significant benefit over only using superset permutations, as the difference between 12-0 and 6-6 ratios is most significant for the relative performance. We attribute this performance boost to the increase in the diversity of the number of people in each scene. The 2-10 ratio, shows a further improvement mainly in absolute pose. 

We then examine the total amount of permutations and show results in the lower part of Tab.~\ref{tab:permutation_ablation}, which shows that increasing the number of permutations improves performance, as expected.
However, the change becomes marginal or even negative with a high number of permutations, as the absolute joint error slightly increases when moving from the 2-10 ratio to the 3-13 ratio, despite a decrease in relative joint error.

\begin{table}[t]
\centering
\resizebox{0.88\columnwidth}{!}{%
\begin{tabular}{@{}cccc@{}}
\toprule
\# Supersets & \# Subsets & \multicolumn{1}{l}{MPJPE$_{\text{rel}}\downarrow$} & \multicolumn{1}{l}{MPJPE$_{\text{abs}}\downarrow$} \\ \midrule
12          & 0         & 58.4                                        & 151.5                                       \\
6           & 6         & 56.5                                        & 150.0                                       \\
2           & 10        & \textbf{56.4}                                        & \textbf{143.0}                                       \\ \midrule
1           & 7         & 56.3                                        & 148.6                                       \\
2           & 10        & 56.4                                        & \textbf{143.0}                                       \\
3           & 13        & \textbf{55.8}                                        & 146.8                                       \\ \bottomrule
\end{tabular}%
}
\caption{Ablation study on the permutation numbers in superset \(n_{\text{sup}}\) and subset \(n_{\text{sub}}\). Upper: we fix the total number of permutations to be 12 and alter the distribution of \(n_{\text{sup}}\) and \(n_{\text{sub}}\).
Lower: we change the total number of permutations while keep the rough ratio of \(n_{\text{sup}}\) and \(n_{\text{sub}}\). All numbers are joint errors in mm. We observe that adding subset permutations and increasing total permutations increases performance significantly.
}
\label{tab:permutation_ablation}
\end{table}

\paragraph{Diffusion Process}
Following \cite{shan_diffusion-based_2023} we use the DDIM diffusion process, allowing us to run inference with only five steps rather than the 1000 steps the model is trained with. As there are arguments from previous works on whether the diffusion process is necessary \cite{zhao2023unleashing}, we evaluate the influence of our diffusion approach on the Haggling dataset. Therefore, we compare \methodname and a version trained and tested without the diffusion process, \ie only the transformer backbone itself. Our results show that integrating the multi-step diffusion process can improve the relative joint error from 75.0 to 55.3 mm and the absolute joint error from 167.5 to 119.5 mm. This shows the necessity of the diffusion process in our model.

\subsection{Qualitative Results on Social Dataset}

\begin{figure}[t]
    \includegraphics[width=\columnwidth]{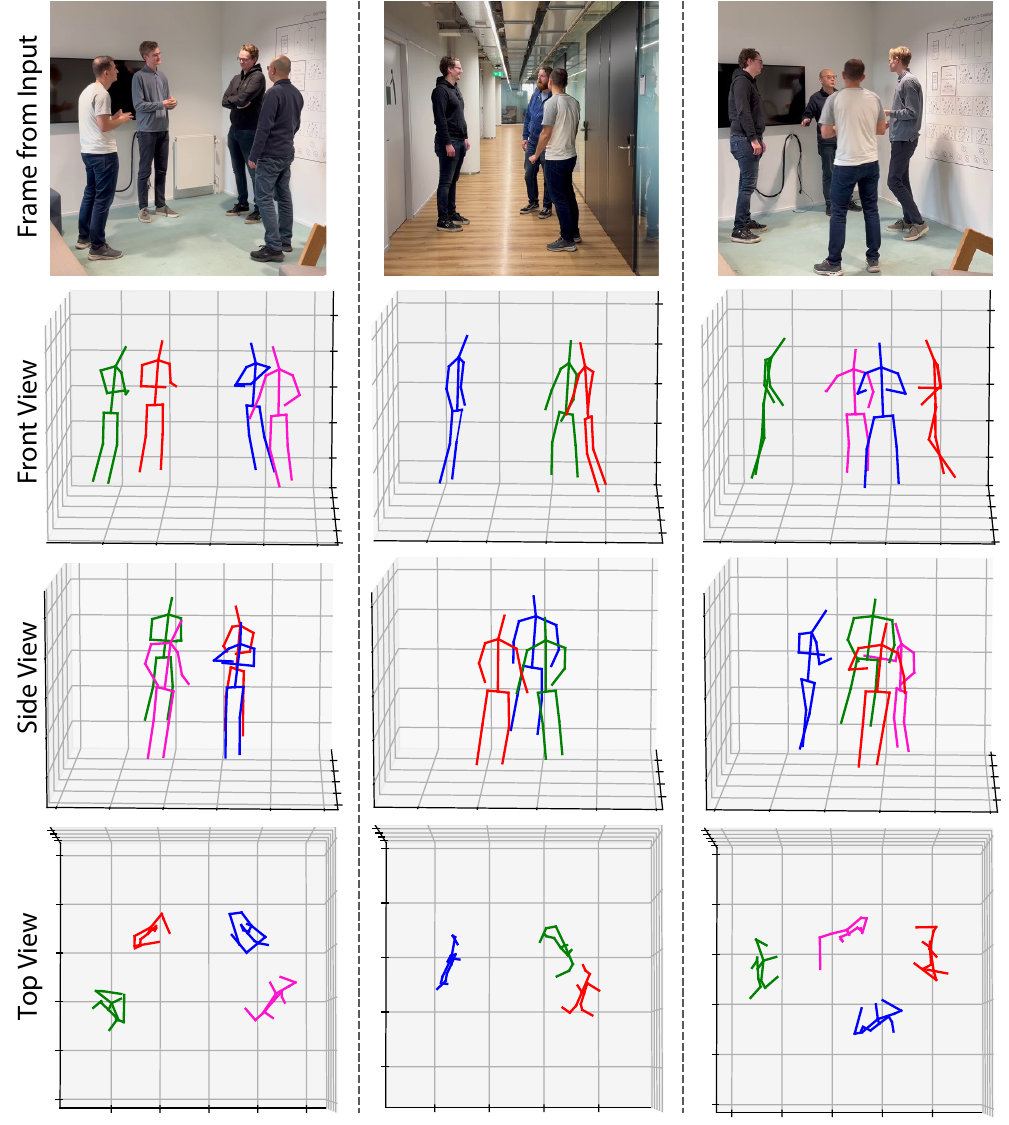}
    \caption{Qualitative results on an in-the-wild setting predicted by \methodname. We show one frame from the video input and corresponding three views of the predicted 3D pose from multiple persons in the scene. Our \methodname demonstrates effective performance in absolute 3D joint prediction, even on highly occluded persons.}
    \label{fig:exp_qualitative}
\end{figure}

To showcase the performance of our method on in-the-wild social interaction, we qualitatively analyze \methodname on in-the-wild social interaction.
The videos are recorded with a camera from a cellphone in different scenes.
We use OpenPose to perform the 2D pose detection on the images, and match persons across frames using the Euclidean distance and Hungarian matching \cite{Kuhn1955Hungarian}. \methodname lifts these 2D detections to the absolute 3D joint positions. As shown in Fig.~\ref{fig:exp_qualitative}, \methodname can successfully locate and place all people from the scene in the 3D world coordinate system, which can be used for social analysis such as F-formations \cite{fiksdal1993conducting}, physical distance between people, and body orientation \cite{varadarajan2018joint,alameda2015analyzing}. Moreover, we observe that even significantly occluded persons are reconstructed well, as can be seen in the second and third columns. The video versions of these results are available in the supplementary material.
\section{Discussion and Conclusion}
\label{sec:conclusion}
\paragraph{Limitations}
Our method is optimized for a specific setting of multi-person interaction and absolute joint estimation. 
In addition, the people in the group interaction must be engaged in a joint activity to form the inter-person relationship that our model can leverage. Lastly, we should have the absolute 3D joint position of each person for training and evaluation.
Given these requirements, there is limited data that we could train and evaluate our model on. 
Specifically, the MuPoTS-3D \cite{mehta_single-shot_2018} dataset that is popularly used in previous works is not suitable for our model. Mainly due to the training data of MuPoTS-3D, which is artificially combined of multi-people from different scenes without any real interaction between them. Additionally, most evaluation scenes of MuPoTS-3D do not contain natural social interaction.

As for future work, we see an opportunity to explore the usage of audio information in group activities for the pose estimation task. Given the fast development of multi-modality learning, we believe it is feasible to add the audio modality to aid the pose inference, given the correlation between speech and body gesture.

\paragraph{Conclusion}
In this paper, we propose a novel 2D-to-3D human pose-lifting method \methodname specifically for the multi-person group interaction setting. To learn the intra- and inter-person relationship, we propose dynamic spatial-temporal attention across all people, person encoding to distinguish each person, and permutation learning to increase the training data diversity. We successfully show the better performance of \methodname compared to the SOTA pose lifting method, and demonstrate its advantages in handling occlusion scenarios. We further showcase the qualitative results of \methodname on in the wild group social interaction data.
{
    \small
    \bibliographystyle{ieeenat_fullname}
    \bibliography{main}
}


\end{document}